# Sensor Response-Time Reduction using Long-Short Term Memory Network Forecasting

Simon J. Ward, Muhamed Baljevic, and Sharon M. Weiss

***Abstract*— The response time of a biosensor is a crucial metric in safety-critical applications such as medical diagnostics where an earlier diagnosis can markedly improve patient outcomes. However, the speed at which a biosensor reaches a final equilibrium state can be limited by poor mass transport and long molecular diffusion timescales that increase the time it takes target molecules to reach the active sensing region. While system and sensor design optimization can promote molecules reaching the sensing element faster, a simpler and complementary approach for response time reduction that is widely applicable across many sensor platforms is using time-series forecasting to predict the steady-state sensor response. Here, we show that ensembles of long short-term memory (LSTM) networks can accurately predict equilibrium biosensor response from limited initial time-dependent biosensor measurements, reducing response time by a mean and median factor of improvement of 18.6 and 5.1, respectively. The ensemble of models also provides simultaneous uncertainty estimation, essential for confident predictions and subsequent safety-related decisions. This approach is demonstrated on real-time experimental data collected by exposing porous silicon biosensors to buffered protein solutions using a multi-channel fluidic cell, automating the measurement of 100 biosensors in parallel. The dramatic improvement in response time achieved using LSTM network ensembles and associated uncertainty quantification opens the door to trustworthy and faster responding biosensors, enabling more rapid medical diagnostics for faster clinical decision making that can lead to improved patient outcomes and healthcare access, as well as quicker identification of toxins in food and the environment.**

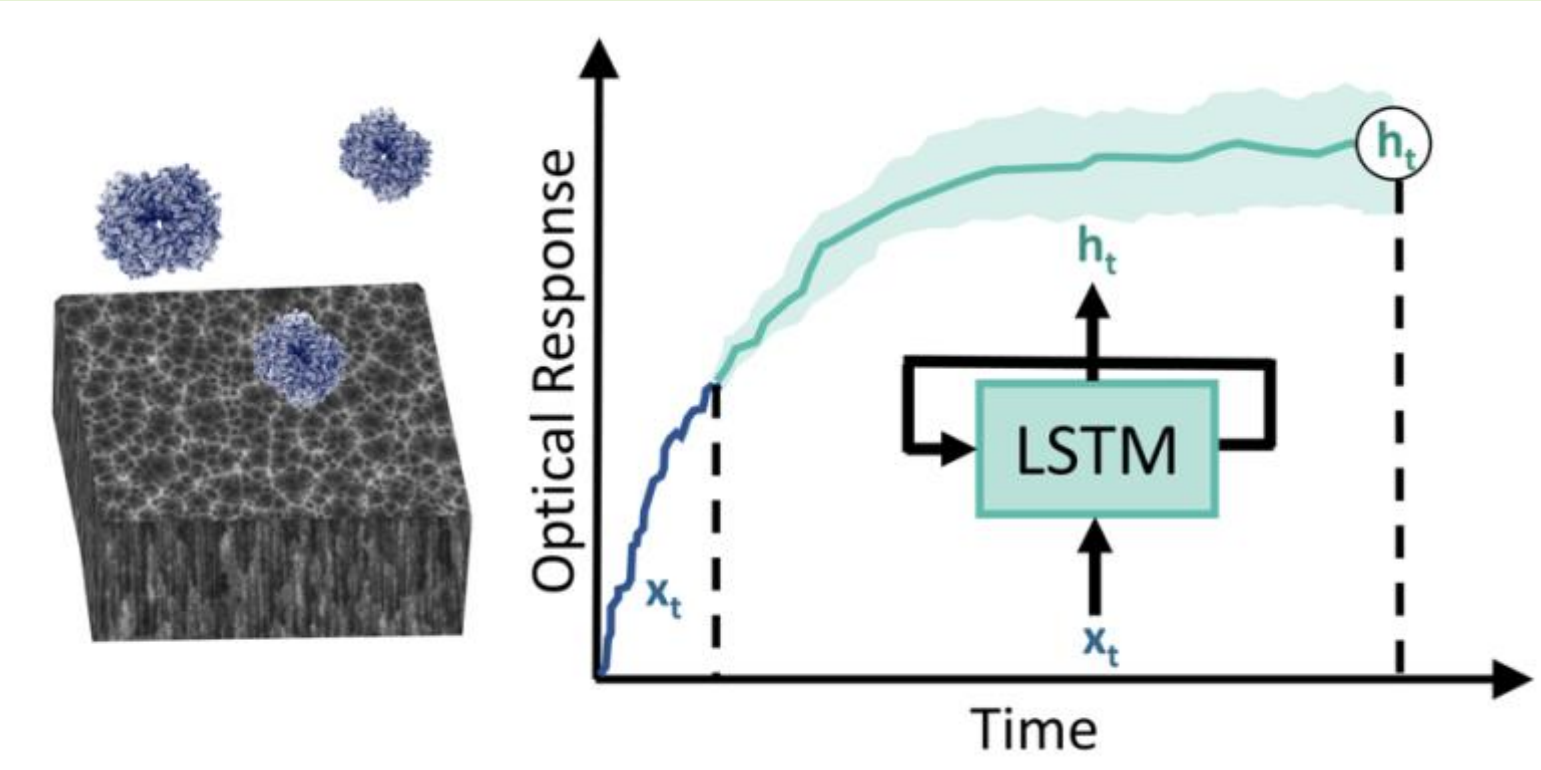

## I. Introduction

DETECTING a target molecule quickly is crucial for a wide variety of safety critical sensing applications, spanning biosensing and chemical sensing. A faster turn-around time for medical diagnostic test results can improve patient outcomes[1], [2], [3], [4], swifter warning of toxins in food or the environment can prevent harm to human health or wider ecosystems[5], [6], [7], [8], and the sooner hazardous gases can be identified the more effectively preventative action can be taken to protect people, assets, and equipment[9], [10], [11]. A good example in clinical practice for which response time is important is symptomatic hyperviscosity syndrome, which represents a medical emergency that requires not only prompt therapeutic intervention[12], but also close immunoglobulin M (IgM) monitoring in these patients thereafter[13]. Rapid turn-around of serum IgM levels is therefore key in the real-time management of these patients and their optimal clinical decision making.

Rapid sensor response times can be inhibited by diffusion and advection as well as binding kinetics, all of which dictate the time it takes for the sensor to reach equilibrium[14], [15], [16]. There are several strategies for improvement of sensor response times, ranging from optimizing sensor architecture to achieve a more direct and less tortuous path from the bulk analyte solution to the sensing element[17], [18], [19], to introducing pumps to increase the analyte flow rate[15], [16]. However, a much simpler and more cost-effective approach is to use machine learning to predict the ultimate equilibrium state of the

Received xx xxx 2025. This work was supported by The Rawlings Foundation, Inc., Myeloma Research Fund, the US National Science Foundation under NSF award number 2037673, and internal Vanderbilt resources. The associate editor coordinating the review of this article and approving it for publication xx. (Corresponding author: Simon J. Ward).

Simon. J. Ward is with the Department of Electrical and Computer Engineering, Vanderbilt University, Nashville, Tennessee 37235, USA (e-mail: simon.j.ward@outlook.com).

Muhamed Baljevic is with the Vanderbilt-Ingram Cancer Center, Vanderbilt University Medical Center, Nashville, Tennessee 37235, USA (e-mail: muhamed.baljevic@vumc.org)

Sharon. M. Weiss is with the Department of Electrical and Computer Engineering, Vanderbilt University, Nashville, Tennessee 37235, USA (e-mail: sharon.weiss@vanderbilt.edu)

sensor based on initial time series response measurements collected over a much shorter timescale. This method requires no change in sensor design or additional components to achieve a higher flow rate, and it can be applied in parallel to such improvements for further gain in response-time reduction. Moreover, machine learning enabled time series forecasting for response time reduction is potentially applicable to a plethora of different systems focused on detecting either biological or chemical analytes, ranging from accurate, large and expensive benchtop instrumentation to cheap, easy to use and portable devices.

Machine learning and signal processing have previously been used to improve sensor response times by extracting features that provide a more expedient measure of sensor response[20], [21], [22], [23], fitting analytical expressions to the incoming time series data[24], [25], or predicting sensor response either dynamically based on the data available or by using a fixed or sliding window[26], [27], [28]. Often filtering is applied in conjunction with these strategies to provide a more stable prediction[22], [24]. However, there are several drawbacks to these approaches. Firstly, they are often sensor-platform specific and require prior knowledge of the shape and form of the sensor response, either an analytical expression to approximate the time-evolution of the response or features that are characteristic of a given sensor architecture and system design. This prerequisite knowledge limits generalizability of the approach since it may not be portable between different sensor types or even across the same sensor platform. Secondly, noise susceptibility is a pervasive challenge, particularly for deconvolution algorithms which amplify noise[24]. Thirdly, in many previous relevant studies, dataset size is limited[20], [23], [24], [26], [27], sometimes consisting of less than 15 experiments because they are expensive to perform, which makes it hard to claim generalizability. Additionally, uncertainty quantification, widely considered an important part of predictive modeling and, consequently, commonly incorporated across many fields[29], has not, to our knowledge, previously been considered in other work aiming to make predictions to decrease sensor response time. Finally, to our knowledge, there has been little application of these techniques to biosensing. Almost all prior work focuses on gas sensing applications despite the fact that the response time of biosensors is typically more than one order of magnitude slower than gas sensors, and thus biosensors have more to be gained by computationally reducing response time.

In this work, we introduce a machine learning-based approach incorporating an ensemble of long short-term memory (LSTM) networks to accurately predict the final response of biosensors, faster than the timescales of equilibrium molecular attachment by a mean factor of 18.6x and median factor of 5.1x. Importantly, the predicted response is provided alongside a quantification of the associated uncertainty. To estimate uncertainty, we followed the approach of aggregating probabilistic outputs of an ensemble of models as detailed elsewhere[30], due to its simple, scalable, well calibrated and high accuracy uncertainty estimation capabilities. Predictions are made dynamically at each time step from real-time measurements. Accordingly, the time taken to return the final result is variable based on prediction uncertainty and stability, so the user is never waiting longer than necessary. If the uncertainty of the predicted response is unacceptably high, the test run-time can be extended until confidence in the predicted result improves sufficiently. However, if the uncertainty estimation remains high after some predetermined threshold time, the test would be stopped and concluded invalid. We elected to use porous silicon (PSi) optical biosensors, a well-established sensing platform which has found success in applications including detection of DNA[31], glucose[32], bacteria[33] and other biological molecules[34], [35], as the model system to demonstrate the utility of our sensor response time reduction approach. While PSi has many advantages for biosensing, including tunable pore size and morphology[36], [37], inexpensive and straightforward fabrication of thin film structures, and large internal surface area presenting a multitude of binding sites for biomolecule adsorption, mass transport is a formidable challenge for PSi biosensors[16], [38]. Hence, demonstrating a method for reducing the response times of PSi biosensors could facilitate the use of PSi biosensors for point-of-care diagnostic devices. Overall, the data-driven approach reported here for reducing the response time of sensors is expected to be portable between sensor assays and platforms because LSTM networks can learn (1) features from data rather than requiring manual feature engineering, (2) the distribution of response curve shape as a function of concentration without theoretical approximations, and (3) how to effectively distinguish signal from noise. As such, applying the same workflow to a different sensor platform likely would only require retraining with representative data from that system, although the resulting efficacy and performance remain to be determined experimentally.

## II. Materials and Methods

All chemicals were of analytical grade and were used without further purification. Boron doped, single side polished, silicon wafers (⟨100⟩, 0.01−0.02 Ω cm) 100 mm in diameter and 500−550 μm thickness, purchased from Pure Wafer, were used to fabricate PSi thin films. De-ionized (DI) water with a resistivity of 15 MΩ cm from a Millipore Elix water purification system was used to wash equipment and samples. HEPES buffer was acquired from Sigma-Aldrich, ethanol and bovine serum albumin (BSA) were obtained from Thermo Fisher Scientific, and aqueous hydrofluoric acid (HF) (48-51%) was purchased from Acros Organics.

### A. Single Layer PSi Biosensor Fabrication

Single layer PSi thin films were fabricated via electrochemical etching of p-type silicon wafers immersed in a HF electrolyte using a platinum mesh cathode, described in detail elsewhere[39], [40]. An Advanced Micromachining Tools (AMMT) MPSB etching system was used for wafer scale PSi etching. First a sacrificial layer was etched using a current density of 60 $mA \cdot cm^{-2}$ for 100 s and then removed through dissolution in 0.1 M aqueous NaOH mixed with ethanol in a ratio 1:4, in order to control surface pore size and remove any surface contaminants. The wafer was washed with ethanol, dried under nitrogen stream, and subsequently etched again using the same conditions as for the sacrificial layer: 60 mA·cm$^{-2}$ for 100 s. The resulting PSi layer had an average pore size of 18.0 ± 0.3 nm, and was 3.63 ± 0.01 μm thick on average, as determined by scanning electron microscopy (SEM) and both top view and cross-sectional image analysis[39] of four PSi samples, two from one etched wafer and two from another etched wafer. The etched wafer was thermally oxidized in a Lindberg/Blue M box furnace at 800°C for 10 min to passivate the surface forming a layer of $SiO_2$, which increases stability, is less prone to corrosion, and is hydrophilic to allow analyte solution to fill the pores and promote infiltration of the target molecule into the pores. Finally, the wafer was diced into 8 × 8 mm$^2$ square samples using a DISCO DAD3220 dicing saw.

## B. Optical Reflectance Measurements

Broadband white light from a Thorlabs SLS201L tungsten-halogen light source was coupled into one arm of a bifurcated optical fiber and focused to a spot size 5 mm in diameter onto each PSi biosensor using an Olympus SPlan 10x microscope lens. Reflected light was collected by the other arm of the bifurcated fiber and coupled into an Ocean Optics USB4000 spectrometer connected via USB to a PC running Ocean Insight Ocean Direct software. The average of 300 consecutive spectra were collected using an integration time of 3.8 ms, resulting in a total measurement time of 1.14 s.

## C. PSi Reflectance Spectra

Because the nanoscale pores comprising PSi are typically smaller than the shortest visible wavelengths of light used for measurement, PSi can often be considered an effective medium described by a single refractive index. The reflectance spectra of PSi single layer thin films contain approximately sinusoidal Fabry-Pérot interference fringes, which are equally spaced as a function of wavenumber, the inverse of wavelength. The frequency of the Fabry-Pérot fringes is equal to the effective optical thickness (EOT) of the thin film, $2nL$, where $n$ is the effective refractive index and L is the physical thickness of the film. As molecules infiltrate and adsorb inside the PSi layer, the volume fraction of air decreases, as it is replaced by higher refractive index biological material, increasing the effective refractive index and, accordingly, EOT of the PSi layer. The corresponding increase in frequency of the fringes can be quantified using the fast Fourier transform (FFT)-based reflective interferometric Fourier transform spectroscopy (RIFTS) method[41]. In this approach, each measured reflection spectrum is first converted from wavelength to wavenumber by inverting the x-axis, then multiplied by a Hann window to enforce periodicity and zero-padded to achieve the desired frequency resolution. A fast Fourier transform is then applied, and the position of the maximum of the dominant peak in the resulting spectrum is identified to determine the fringe frequency. In this way, the frequency of the interference fringes can be tracked over time, producing a time series that reflects the progressive capture of biomolecules within the Psi layer, and corresponding increase in refractive index.

## D. Real-Time Sensor Response Data Collection

A large experimental throughput is necessary to enable collection of sufficiently large datasets to allow analysis by machine learning algorithms. Typically, real time optical measurements of PSi are collected one sensor at a time in a single channel fluidic cell, or occasionally a fluidic cell with multiple channels has been used to measure a handful (<5) of locations on a PSi sensor[42], [43]. However, in order to increase experimental throughput by many orders of magnitude, we present the novel design of a multiplexed fluidic cell accommodating a large number of PSi sensors which are measured in an automated and sequential manner using an addressable stepper motor to incrementally step between samples. The multi-channel fluidic cell was designed to house up to one hundred individual 8 × 8 mm$^2$ PSi biosensors. The fluidic cell consisted of two plexiglass sheets 3.175 mm thick, laser-cut into disks 260 mm in diameter with several screw holes to hold the sheets together, and guide holes to align them correctly. An inlet and outlet for each channel of the fluidic cell were also laser-cut to have a diameter of 1 mm. For each channel, the silicon substrate side of a PSi sample was secured to the lower plexiglass sheet using double sided scotch tape. O-rings 8 mm in diameter and with 1 mm thickness were placed over each PSi sensor. The upper plexiglass sheet was then guided onto the lower assembly, ensuring all O-rings lay within the boundaries of each PSi sensor and encompassed both the inlet and outlet of the upper plexiglass sheet to create an effective seal. The plexiglass sheets were screwed together along with a calibrated spacer to achieve consistent O-ring compression, isolating the flow paths over each sensor. Together with the physical separation between sensors and rigorous ethanol and water cleaning between experiments, this configuration eliminates cross-contamination. Once secured with screws, the multi-channel fluidic cell was affixed to a Polulu 1206 stepper motor using the mounting hub. Python scripts were used to automate and schedule both the setting of stepper motor position via the Tic T500 USB stepper motor controller board and yaml library, and processing and saving reflectance spectra using the Ocean Direct API. We note that it is imperative for the stepper motor to return to the same position with high accuracy for every sensor to prevent any non-uniformity of PSi film optical thickness from introducing error into the measurements; however, we found that even an inexpensive, low resolution stepper motor was able to return to the same position with micron precision, which provided sufficient repeatability and negligible error. Further details can be found in Appendix A. To the best of our knowledge, this is the first experimental protocol in which real-time measurements of numerous PSi sensors are collected in parallel.

## E. Experimental Procedure

Thirty PSi samples from an etched and diced wafer were washed using ethanol and dried under nitrogen stream before being secured into the multichannel fluidic cell. We note that the total capacity of the fluidic cell was 100 sensors, but there is a tradeoff between the time taken for each measurement in addition to the time to step between sensors, and the time resolution of consecutive measurements. Further speed optimizations would further increase experimental throughput. A dark spectrum corresponding to 0% reflectance measured without illumination from the light source and a reference spectrum indicating 100% reflectance measured with an unprotected gold mirror were collected to calibrate the reflectance measurements. A vial of 200 µL of BSA solution was prepared at one of 13 different concentrations: 40 mg/mL, 20 mg/mL, 10 mg/mL, 4 mg/mL, 2 mg/mL, 1 mg/mL, 0.4 mg/mL, 0.2 mg/mL, 0.1 mg/mL, 0.04 mg/mL, 0.02 mg/L, 0.002 mg/mL, and a control solution consisting of 100% HEPES buffer. The optimum stepper motor positions, as close as possible to the center of each PSi sample, were determined and hard coded into the python control software. The measurement cycle was initiated, which first measured each sensor in turn before application of analyte solution. On the second rotation, the control software paused for an additional 10s over every PSi sensor channel to allow the injection of 25 µL protein solution into the inlet of each fluidic channel in turn. Subsequent measurement loops continued for 250 measurements of each sensor, equivalent to an experiment length of 13 hours. The entire measurement setup, including optical fiber, reflectance probe, stepper motor and multi-channel fluidic cell were encased in an airtight container, with a beaker of water also enclosed to maintain a high humidity locally, which prevented BSA solution evaporation over the extended experiment runtime. For practical deployment, however, evaporation is less of a concern. Unlike the high-throughput configuration used here, a single sensing element can be enclosed in a compact, sealed holder with relative ease. In addition, the operational measurement time for the trained model is only a small fraction of the 13-hour window used to generate the training set, resulting in minor evaporation which would have negligible influence on the measured sensor response. Hence, evaporation is not expected to present a practical limitation for field deployment of the sensor and associated predictive model.

The time series of reflectance spectra for each PSi sensor were analyzed using the RIFTS method, and normalized according to the following equation: $\frac{EOT_n - EOT_0}{EOT_0}$ where $EOT_0$ and $EOT_n$ are the effective optical thicknesses of the PSi at the first timestep and the nth timestep, respectively. In total, the dataset consisted of 387 experimental sensor response examples spread across the 13 concentrations.

## F. Uncertainty Quantification

Of the numerous ways to quantify uncertainty associated with machine learning model predictions, including Bayesian neural networks[44] and Markov chain Monte Carlo (MCMC)[45], we follow the methodology outlined in Ref 30 to estimate uncertainty of the predictions. This work established that a particularly simple, scalable and effective way of estimating uncertainty for deep neural networks was to assume the output of the network is normally distributed, parameterized by predicted mean (µ) and variance ($\sigma^2$) which are captured by two nodes in the output layer. An ensemble of such neural networks was each individually trained by minimizing the negative log likelihood, and the mean and variance predictions of each of these base learners was aggregated to give a more accurate and better calibrated predicted uncertainty. In this study, these principles were applied to LSTM networks. The predictions of µ and $\sigma^2$ were averaged for each of the LSTM network base learners, taking into account both within class and between class variance, yielding a point prediction indicated by mean µ and a measure of uncertainty or prediction spread indicated by variance $\sigma^2$.

Rather than varying the architecture of the base learners, using Monte-Carlo (MC) dropout, or bootstrapping the dataset, we employed deep ensembles with diversity induced solely through random parameter initializations and data shuffling. This approach avoids the potential performance degradation associated with bootstrapping, where each model trains on only 63% unique data points, and has been shown empirically to often significantly outperform, or at least match MC-dropout methods in terms of predictive accuracy and uncertainty calibration[30]. While MC-dropout offers computational advantages by requiring only a single trained model, deep ensembles provide superior performance for this application where accurate equilibrium prediction is critical.

## G. Data Analysis

LSTM networks, a class of recurrent neural networks (RNN), have been widely adopted for time series forecasting and anomaly detection across diverse domains, including machine health[46], speech enhancement[47], weather forecasting[48], and text classification[49]. LSTMs were designed to mitigate the problem of exploding and vanishing gradients experienced by traditional RNNs, significantly improving the long-term memory of previous time points in sequential data[50]. This is achieved by enhancing gradient flow using a LSTM internal memory which runs through each sequential LSTM cell, and is maintained with three gates. The first is the input gate, which determines the extent to which the current input to the LSTM cell should be stored in the internal cell memory, the second is the output gate which calculates the proportion of internal cell memory that should be included in the output of the cell, and the third is the forget gate which identifies less relevant information currently stored in the internal cell memory which should be forgotten. As a result of this architecture, LSTM networks are well suited for the rapid prediction of equilibrium sensor response due to their ability to learn features without requiring manual feature selection, to learn to distinguish signal from noise, and to learn long and short-term dependencies in sequential data, all of which promote generalizability.

The primary baseline considered in this work is the experimentally measured response time of the sensor without prediction, which reflects the fundamental limitation imposed by mass transport and molecular diffusion, and time required to reach a confident and actionable result. All reported performance gains are therefore quantified in terms of reduction in this experimentally observed response time.

TABLE I
LSTM MODEL ENSEMBLE HYPERPARAMETERS

| | |
|---|---|
| Sequence Length | 250 |
| Train/Validation/Test split | 60%/20%/20% |
| Splitting Strategy | Stratified by concentration |
| Normalization Method | Global min-max scaling |
| Normalization Range | [0 ,1] |
| Ensemble Size | 15 base learners |
| Number of LSTM layers | 3 |
| Layer 1 LSTM blocks | 50 |
| Layer 2 LSTM blocks | 500 |
| Layer 3 LSTM blocks | 2 |
| Output Layer Activation | Softplus (log(1 + exp(·)) |
| Variance Numerical Stability Constant | $10^{-6}$ |
| Optimizer | Adam |
| Optimizer Learning Rate | 0.001 |
| Batch Size | 4 |
| Number of Epochs | 1,500 |
| Best Model Selection | Minimum validation loss |
| Framework | TensorFlow/Keras |
| Numerical Precision | FP32 |

Summary of the principal hyperparameters defining data preprocessing, LSTM model architecture, loss function, optimizer settings, and ensemble training procedure.

Models were implemented in TensorFlow (version 2.15.1) using the keras API (version 2.15.0)[51] and built-in LSTM layers. Each LSTM layer was configured to return a sequence of 250 outputs, one for each input time step. The learning target for each input sequence is the final observed sensor response, replicated across the sequence length. Each output prediction in the sequence is made having only seen data from the current and past timesteps, so will typically become increasingly more accurate as the sequence goes on and more data from the input sequence is seen by the model.

The base learner architecture, informed via limited hyperparameter tuning on the validation set, comprises 50 input neurons, one hidden layer with 500 units, and two output neurons corresponding to the predicted mean and variance.

To obtain both point predictions and calibrated uncertainty estimates, ensembles of 15 independent base learners were trained[52] by treating the observed equilibrium sensor response as a sample from a (heteroscedastic) Gaussian distribution with the predicted mean and variance[30], and minimizing the negative log likelihood. For a given training example $(y_n, x_n)$, the loss is therefore defined as:

$$-\log p_\theta(y_n|x_n) = log\frac{\sigma_\theta^2(x_n)}{2} + \frac{(y_n - \mu_\theta(x_n))^2}{2\sigma_\theta^2(x_n)} + const \qquad (1)$$

where $p_\theta(y_n|x_n)$ denotes the likelihood of the target $y_n$ given the input $x_n$ under model parameters $\theta$. The network outputs $\mu_\theta(x_n)$ and $\sigma_\theta^2(x_n)$ correspond to the predicted mean and variance of a Gaussian distribution, respectively.

To ensure the output of the model is always positive (necessary for both the variance and mean given the target data are normalized to the range [0,1]), a softplus activation function (log(1 + exp(·))) is applied to both mean and variance outputs, and an additional small numerical stability constant $\varepsilon = 10^{-6}$ is added to $\sigma_\theta^2(x_n)$. Ensembles were used to increase accuracy and prediction stability, and for better calibrated uncertainty quantification[30], [52].

The data was first randomly shuffled and split by sensor ID into train, validation and test sets at a ratio of 3:1:1, stratified by BSA concentration. The maximum and minimum sensor response values across all time steps and all examples in the training set were used for min-max scaling to normalize the train, validation and test sets, to avoid data leakage. All models were trained using the Adam optimizer[53] with a constant learning rate of 0.001, a batch size of 4, and 32-bit floating point precision. Training was carried out for 1,500 epochs on a Nvidia RTX 3080 GPU. A summary of all hyperparameters is provided in Table 1.

## III. RESULTS AND DISCUSSION

The achievable reduction in response time using the predictions from an ensemble of LSTM networks was investigated using the experimental dataset measured by exposing buffered BSA solutions to PSi biosensors and comparing the initial intrinsic experimental sensor response time to the response time of the model's predictions.

### A. Experimental Dataset

The experimental dataset of real-time optical reflectance measurements collected by exposing BSA in HEPES buffer at one of 13 different concentrations to PSi biosensors is shown in Fig. 1. PSi reflectance spectra are analyzed using the RIFTS method and expressed in terms of the fractional change in EOT, which allows comparison between PSi sensors with slightly different baseline EOT. These variations are due in part to the random sampling of the distribution of pore size and morphology, and small spatial variations in PSi. It is clear from Fig. 1 that even when concentration is held constant there is large variability in equilibrium response, attributed primarily to the highly variable interactions between molecules and the internal PSi surface rather than variation in PSi sensor properties. In Fig. 1 (b), this is manifested by the range of final equilibrium responses for a single concentration, illustrated by a single color, and in Fig. 1 (c) this is captured in the sizeable error bars which represent the standard deviation of equilibrium responses at each concentration. We note that in Fig. 1 (c), the Redlich Peterson adsorption isotherm[54], representing imperfect monolayer adsorption and previously found to best fit the data[55], provides general agreement with the data showing how the equilibrium fractional EOT change depends on BSA concentration. For the purposes of this study, the variation in equilibrium response for a given concentration is not a concern; in fact, it ensures the equilibrium values in the training data are more uniformly distributed. Additionally, although this study focused on PSi sensors and BSA as a model analyte, the approach of using LSTM networks is, in principle, generalizable and portable to other biosensing platforms that produce time-dependent response signals. This generalizability arises because the network learns temporal features directly from experimental data, rather than relying on sensor-specific kinetic models or handcrafted features. Consequently, the same modeling framework could be applied to higher affinity assays that rely on specific binding, and for which response curves generated using the same concentration of target molecules would be much more closely grouped, mitigating this issue. We note that the ensemble of LSTM models predicts the equilibrium sensor response from early temporal measurements, decoupled from concentration calibration, which is a distinct and addressable downstream issue.

While this BSA assay has long response times (mean of 3.95 hours), which even in the context of PSi biosensors and other platforms is relatively slow, the focus of this work was the relative reduction in experimental response time that predictions from LSTM network ensembles can provide, rather than the absolute value. Simple steps to improve response time for this assay, in addition to the predictive time-series modeling approach reported here, include flowing analyte solution across the sensors rather than using a configuration without analyte flow, decreasing the thickness of the PSi films, and increasing the pore size, all of which promote faster mass transport.

Fig. 1 also demonstrates the non-linearity of the complex molecular adsorption and diffusion dynamics of BSA in the pores, leading to response curve with a shape that depends on concentration, and cannot be accurately fit with a simple analytical expression, which may be in part a consequence of the non-uniformity of the size and morphology of the pores across a given silicon wafer. For BSA concentrations below 0.2 mg/mL, the signal to noise ratio (S/N) is particularly low and there is no systematic difference in the final equilibrium response. In contrast, for concentrations between 0.2 mg/ml and 2 mg/ml, the equilibrium response is highly correlated with increasing concentration and S/N also rises accordingly. Finally, at concentrations above 10 mg/ml, the equilibrium response plateaus as all adsorption sites in the pores are filled, and the PSi sensors are saturated.

The reduced signal quality at low BSA concentrations (<0.2 mg/mL) reflects inherent PSi instability in aqueous buffer[56]. Although thermal oxidation at 800°C substantially improves stability, the oxidized surface remains modestly susceptible to degradation over the 13-hour experimental timescale through hydrolysis of Si-O bonds and dissolution into orthosilicic acid in buffered solutions[57]. At low analyte concentrations, this degradation introduces baseline drift and noise. Conversely, at higher protein concentrations, the layer of surface adsorbed BSA molecules acts as a passivating biofilm, reducing further degradation and improving signal stability[58].

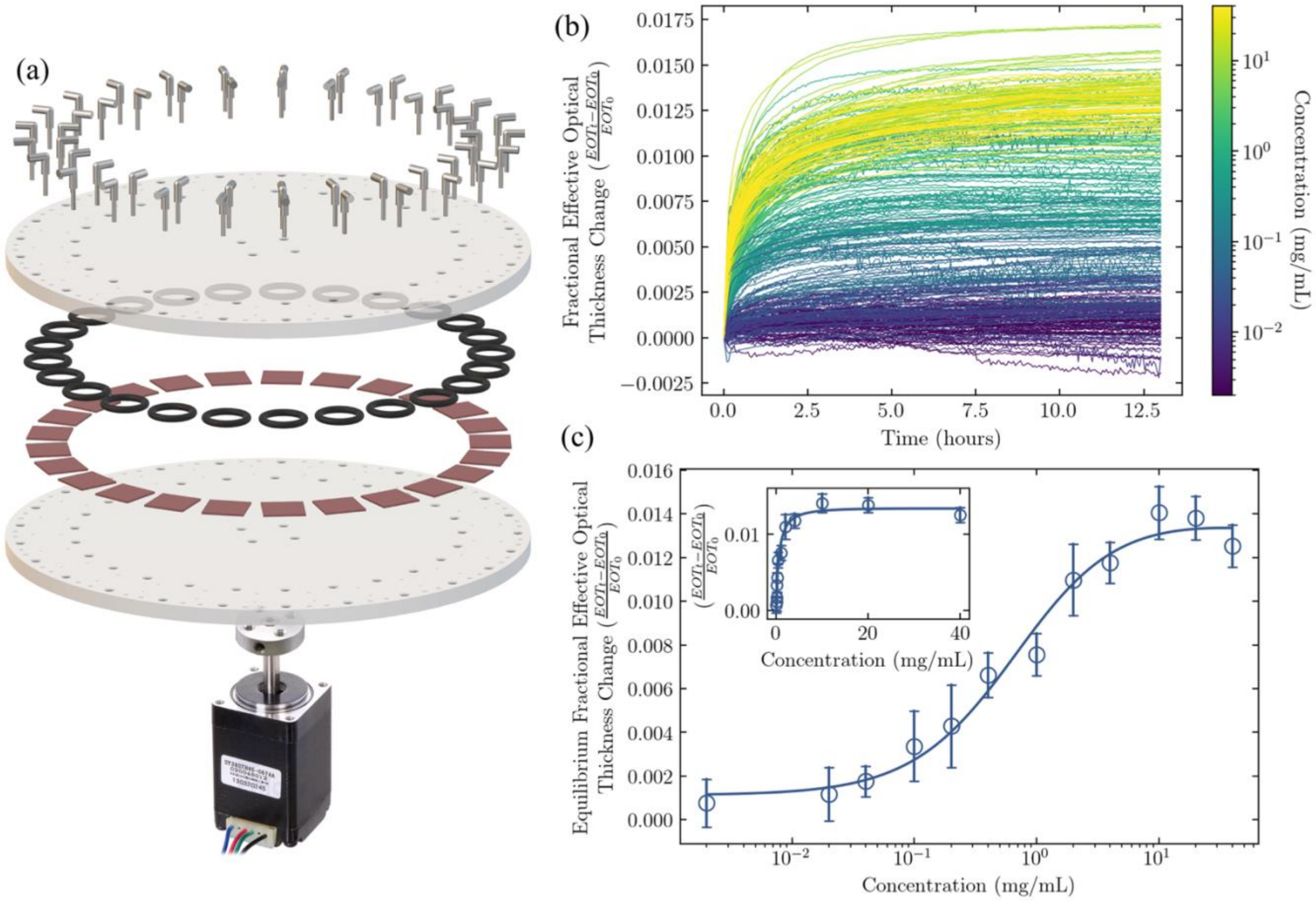

Fig. 1. (a) Diagram of the multi-channel fluidic cell for high throughput real-time data collected by exposing single layer PSi sensors to solutions of BSA in HEPES buffer, including the stepper motor, mounting hub, bottom laser cut plexiglass sheet, PSi sensors, o-rings, top laser cut plexiglass sheet, and inlet and outlet tubing. The experimental dataset expressed in terms of the fractional change in EOT, is shown in both (b) the raw temporal sensor response curves, and (c) adsorption isotherm fit with the Redlich Peterson model[54] on a log scale. The inset of (c) shows the same data on a linear scale, and the error bars indicate the standard deviation of between 26 and 40 measurements for a single concentration.

### B. Evaluation of Model Performance

Several representative examples of sensor response curves from the previously unseen test set alongside the model predictions are shown in Fig. 2. The ideal predictive model response is shown with the dashed line, which initially lies at half the normalized range of sensor responses, and instantaneously jumps to the target equilibrium response, with an infinitesimally small response time. The uncertainty of model predictions is illustrated by the blue shaded regions, spanning two standard deviations on either side of the mean prediction of the ensemble, encompassing 95% of the predicted distribution. The model's predictive ability is shown through the quick convergence to the equilibrium response value – making predictions on relatively few initial time series data points and with low variance–for examples with relatively high S/N (Fig. 2 (a), (b), and (c)). These high S/N examples are typically obtained using higher concentrations. In other cases, particularly for response curves with low S/N which correlates with low BSA concentration, the model predictions take a lot longer and a lot more time series data to converge, sometimes never stabilizing at the correct equilibrium response (Fig. 2 (d), (e), and (f)). Noise sources which reduce S/N include drift in temperature, humidity, as well as vibrations. While for this work it was advantageous to have a range of S/N examples in the experimental dataset to test the limits of the model, these noise signatures could be drastically reduced when using an assay with antifouling layers and stronger binding affinity. Furthermore, the degraded model performance due to poor S/N was reassuringly accompanied by increased prediction uncertainty indicated by the shaded regions in Fig. 2. The uncertainty is always initially high as the model is making predictions on a very small number of data points, but in most cases decreases over time as more data is available and predictions become more accurate.

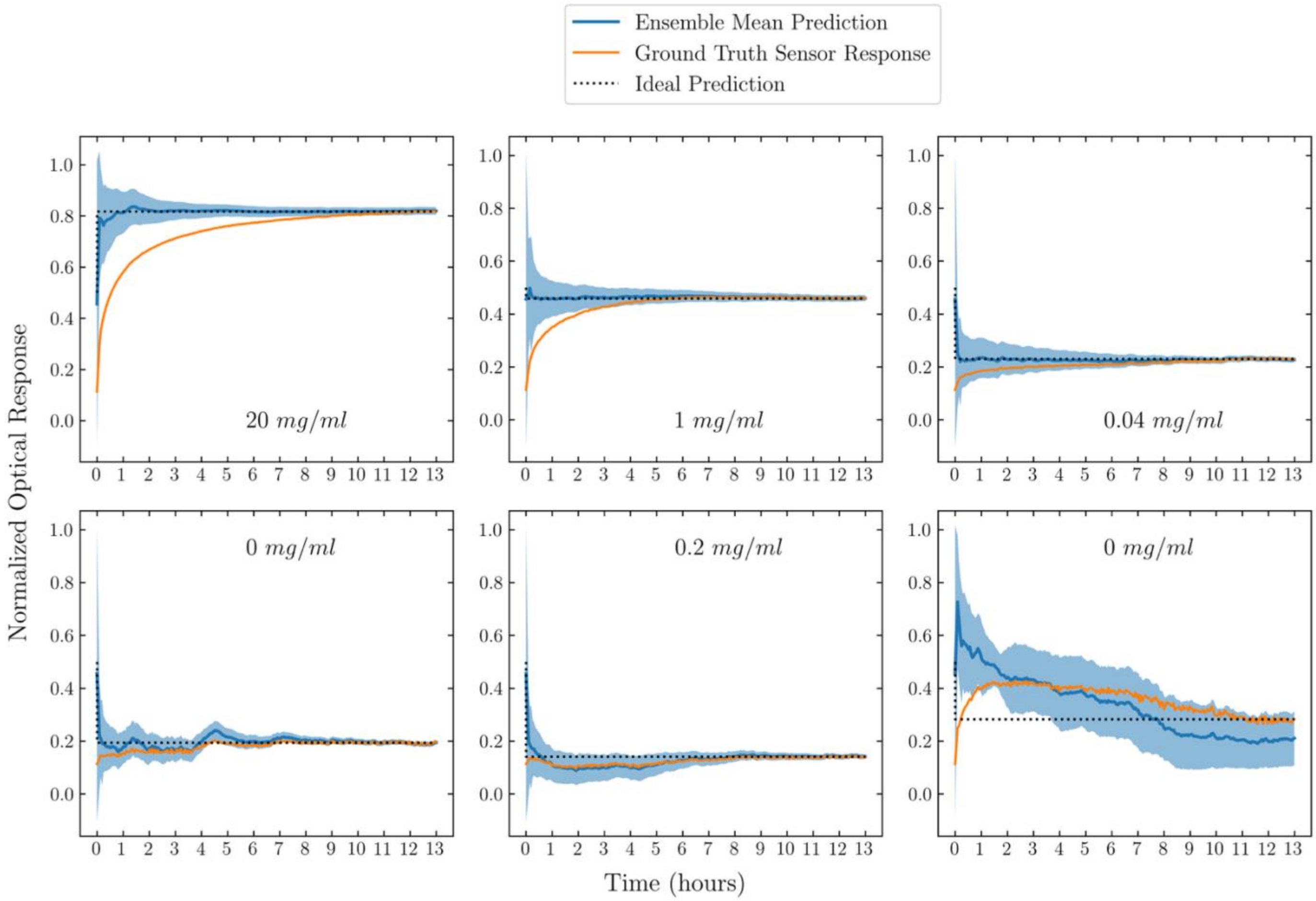

Fig. 2. Series of examples of PSi biosensor time-series responses from the test dataset (orange) which are previously unseen by the model, and corresponding model predictions (dark blue) at every timestep using only data from the current and all previous timesteps. The shaded blue region is a measure of the predicted variance, or uncertainty, in the equilibrium response prediction, encompassing two standard deviations on either side of the ensemble mean, capturing 95% of the distribution. The black dotted line is the ideal model prediction, which is initially a normalized response of ~0.5 at t=0 when the model has no information about the sensor response, then instantly converges to the equilibrium sensor response. The upper three examples in the panel exhibit a much higher S/N than the lower three examples, and, accordingly, superior model performance can be observed in each of the upper three examples in the panel.

When deploying this model, an acceptable threshold of both uncertainty and prediction stability would be chosen; the decision of when to return the test result could then be made dynamically according to when the model output is within these thresholds. This results in a test time that is never longer than necessary, which is not the case when a predetermined fixed window of data is used for prediction. Additionally, a maximum response time cutoff could be defined; sensor responses, for which uncertainty or response instability remained high (above the aforementioned acceptable threshold) beyond the maximum cutoff time, would be deemed invalid, and the test rerun.

In order to determine the best size of the ensemble to use for this application, the number of base learners was varied and the accuracy of ensemble predictions was evaluated on the validation dataset, which was not explicitly used during training. The ensemble size chosen was 15 base learners: increasing ensemble size beyond that offered negligible improvement in response time and only served to increase the complexity and computational burden of the models. Further details can be found in the appendix (Fig. B1).

The standard measure of the speed of response of a sensor is the $T_{90}$ time, which is defined as the length of time taken for a sensor response to reach 90% of its ultimate equilibrium value. This metric implicitly assumes a monotonic approach to equilibrium, as is typical for physical sensors.

In contrast, machine learning–based predictions of sensor response are not constrained to be monotonic and may transiently overshoot or oscillate around the equilibrium value due to experimental noise and initially limited input data. In such cases, the classical $T_{90}$ definition based on a one-sided threshold crossing can underestimate the effective response time by triggering during a temporary overshoot.

To address this, the $T_{90}$ definition was adapted in this work to quantify response stabilization rather than threshold crossing. Specifically, the response time was defined as the earliest time at which the predicted sensor response enters and remains within ±10% of the final equilibrium response relative to the baseline. That is, the response is considered complete only once the prediction has settled within a bounded region around the final equilibrium value.

Formally, let $y(t)$ be the predicted sensor response, $y_0$ the baseline response, and $y_{eq}$ the equilibrium response estimated from long-time measurements. We define the overshoot-aware response time, denoted $T_{90}^{\pm}$, as the earliest time $t$ such that:

$$\forall \tau \geq t, \qquad \left|y(\tau) - y_{eq}\right| \leq 0.1\left|y_{eq} - y_0\right| \tag{2}$$

That is, $T_{90}^{\pm}$ is the time at which the predicted response enters and remains within a ±10% band around the equilibrium response.

For monotonically increasing sensor responses, $T_{90}^{\pm}$ is equivalent to the traditional $T_{90}$ definition, as the response enters the ±10% tolerance band only once upon crossing the 90% threshold and remains there thereafter, yielding identical response times. Differences between the metrics arise only for non-monotonic responses, where classical $T_{90}$ can be biased toward earlier times by transient overshoot, triggering on a temporary threshold crossing that does not reflect true equilibrium. By requiring the predicted response toremain within a bounded region around equilibrium, $T_{90}^{\pm}$ is robust to such overshoot. In the presence of low S/N, classical $T_{90}$ may be triggered prematurely by stochastic fluctuations which cross the 90% threshold, while conversely $T_{90}^{\pm}$ may be prolonged if noise prevents settling within the tolerance band. Similarly, drift in the sensor response can alter the apparent equilibrium value, delaying either criterion being met, or for the classical $T_{90}$, potentially shortening the estimated response time by inducing an early threshold crossing.

A comparison of the experimental response time and response time resulting from the predictions of an ensemble of 15 LSTM networks is shown in Fig. 3. Specifically, the lower part of Fig. 3 (a) shows a histogram illustrating the distribution of $T_{90}^{\pm}$response times for the original experimental dataset and the LSTM ensemble model predictions. The improvement in response time afforded by the model predictions is demonstrated through the large shift in the overall distribution to lower response times. However, there are still some prediction-based response times that remain high, which can be understood with reference to the normalized variance plot in the upper part of Fig. 3 (a). The normalized variance for each experimental sensor response curve is $\frac{\overline{\sigma^2}}{Response(t=13\ hrs)}$ where $\overline{\sigma^2}$ is the average predicted variance, or uncertainty, for every time point in a given sensor response, and $Response(t = 13\ hrs)$ is the corresponding equilibrium sensor response at the end of the experiment. Normalized variance is therefore a measure of the model's confidence (correlated to S/N); the overall linearly increasing trend in the upper part of Fig. 3 (a) indicates that when the model performed poorly this was accompanied by increased uncertainty the majority of the time. An alternative representation of the distribution of factor of improvement in response time provided by the LSTM network ensemble model is shown in Fig. 3 (b, c). The ratio of original experimental $T_{90}^{\pm}$ time to model prediction $T_{90}^{\pm}$ time is presented in the form of a box and whisker plot. The summary statistics reporting the mean and median factor of improvement of 18.6 and 5.1 respectively, are also reported. We note that the lower quartile of the distribution of improvement in $T_{90}^{\pm}$ is 1.0, indicating that a subset of sensor responses exhibit little or no reduction in the predicted $T_{90}^{\pm}$. As shown in Fig. 3 (a), longer predicted $T_{90}^{\pm}$ times are accompanied by increased relative predictive uncertainty, a behavior that is most pronounced for sensor responses with low S/N. In these cases, the ensemble prediction typically does not sufficiently stabilize and the prediction uncertainty remains high. For field deployment, it is therefore straightforward to define thresholds on both uncertainty magnitude and prediction stabilization that would identify this and classify such tests as invalid. This determination, which would lead to test repetition under both conventional and machine-learning-assisted workflows, could be made substantially earlier than the full experimental run time, thereby still reducing overall assay time by allowing early termination of uninformative or unreliable tests. Additionally, since the lower-quartile outcome primarily reflects limitations in sensor signal quality, further improvements could be made through sensor-level enhancements that improve S/N characteristics.

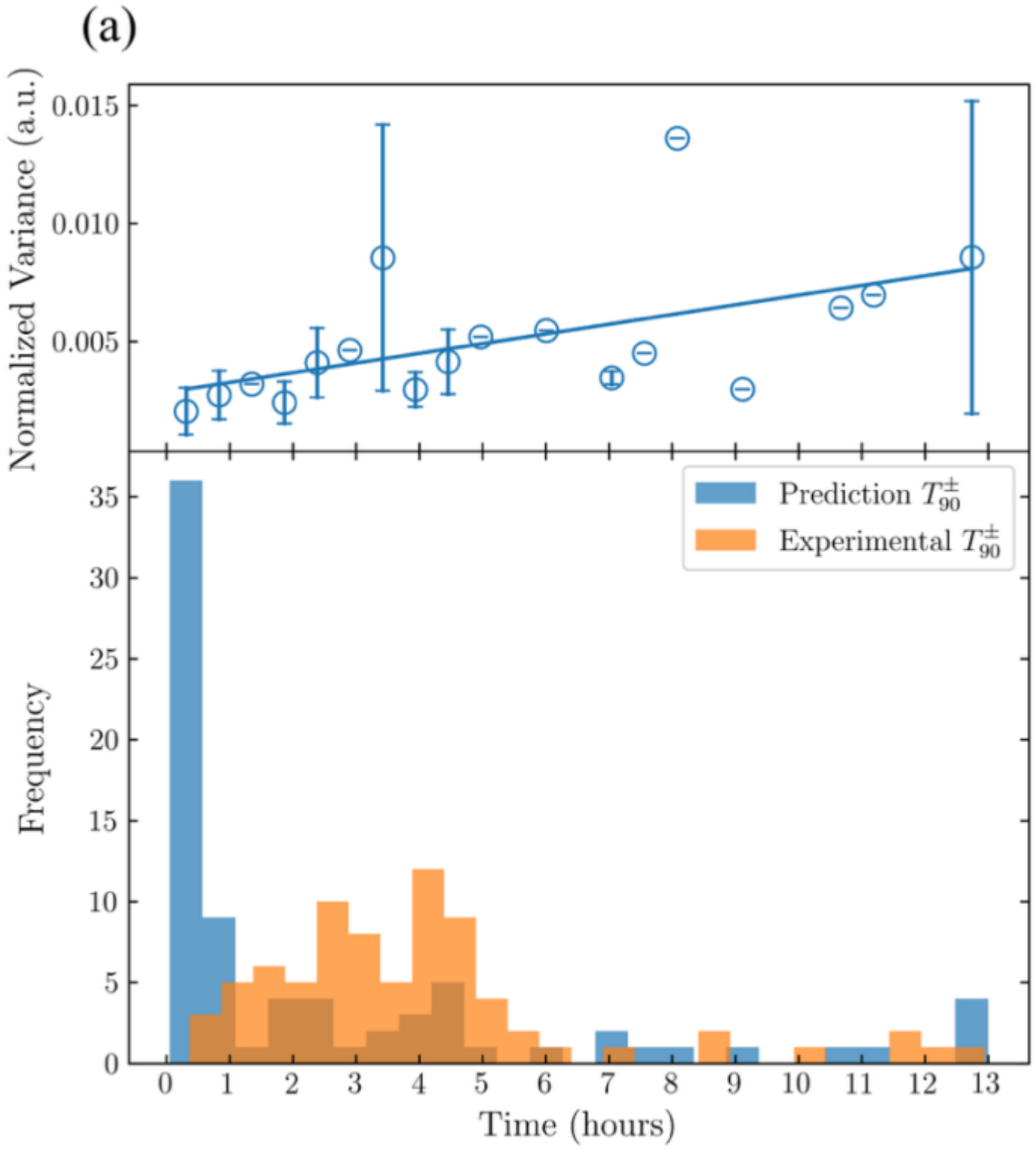

(b)

Descriptive Statistics of the Ratio of Experimental to Predicted $T_{90}^{\pm}$ Response time

| | |
|---|---|
| Test Set Size | 78 examples |
| Mean | 18.6 |
| Lower Quartile | 1.0 |
| Median | 5.1 |
| Upper Quartile | 26.3 |

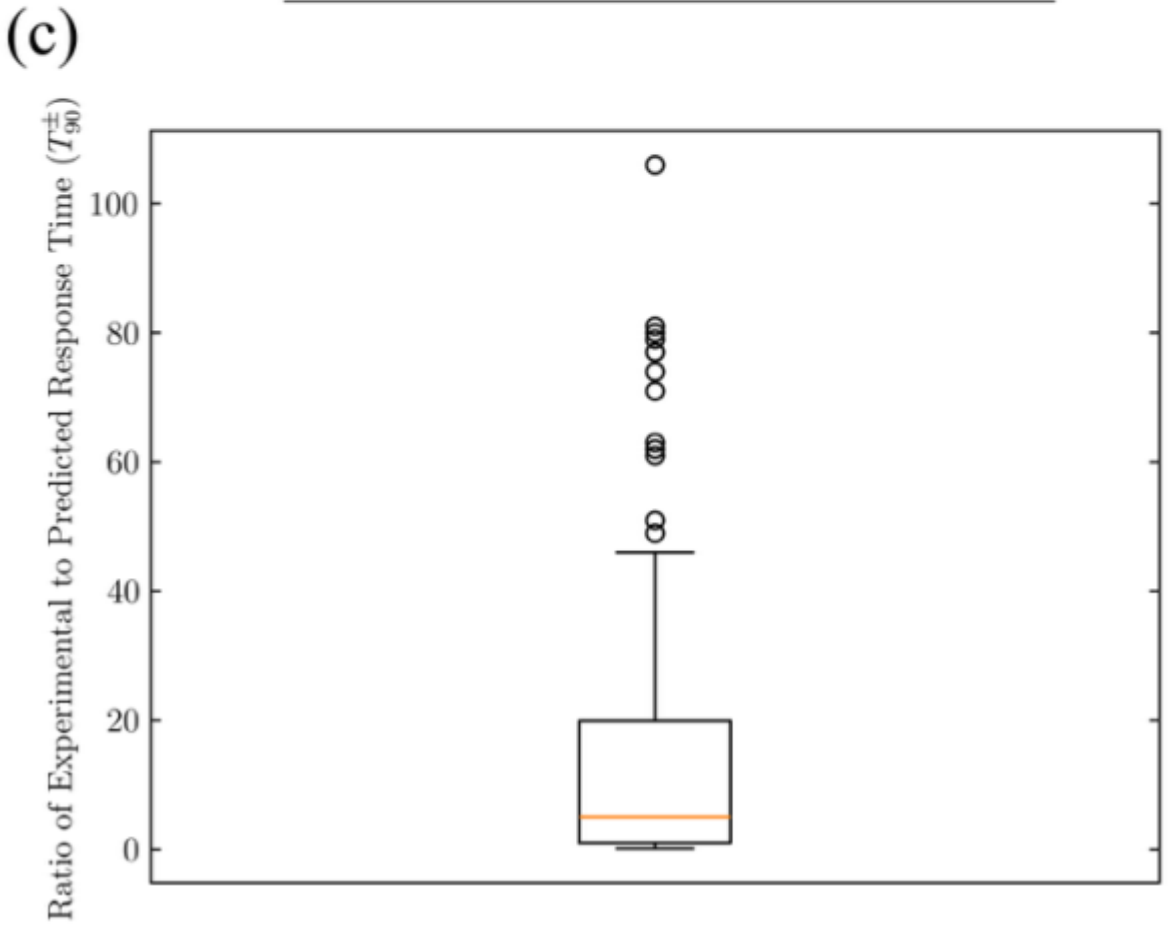

Fig. 3. Comparison of $T_{90}^{\pm}$ response times of the unprocessed experimental data and after application of the model composed of an ensemble of LSTM networks, illustrated by (a) a histogram showing the two distributions of $T_{90}^{\pm}$ response times, and the ratio of experimental to model prediction $T_{90}^{\pm}$ response time displayed in the form of (b) descriptive statistics and (c) a box and whisker plot. The higher this ratio, the greater the factor of improvement. Additionally, (a) includes the time averaged predicted variance, or uncertainty, of the ensemble of LSTM networks for each sensor response in each bin of the histogram. This average variance is normalized by the equilibrium value of the sensor response, indicative of S/N.

We next compare LSTM performance against traditional RNN and gated recurrent unit (GRU) networks, trained and evaluated under identical conditions, as a benchmark. Fig. 4 compares the response time improvement achieved using LSTM, GRU, and traditional RNN models as a function of analyte concentration, quantified by both the mean and median reduction in $T_{90}^{\pm}$ relative to the experimentally measured non-prediction-aided response time. All three model architectures yield substantial reductions in response time. However, LSTM ensembles consistently outperform both GRU and RNN models across the full concentration range and for both statistical measures. The superior performance of LSTM models is consistent with their enhanced capacity to retain and manipulate long-range temporal information via an explicit memory cell, enabling the integration of informative early-time dynamics while attenuating noise. This capability is particularly well suited to non-stationary, monotonic biosensor responses with unknown asymptotic values, where accurate prediction requires inferring a latent equilibrium state from limited early-time observations.

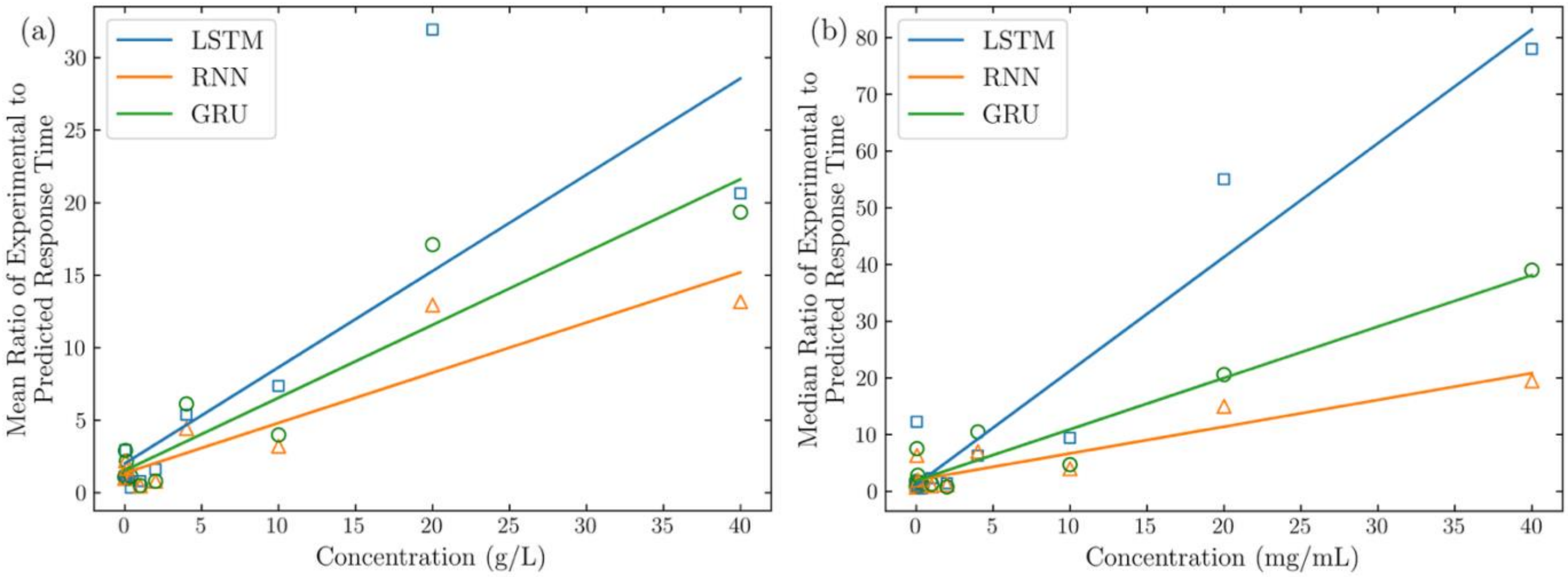

Fig. 4. Comparison of response time reduction achieved on test set using LSTM, GRU, and traditional RNN models, as a function of analyte concentration. Specifically, (a) mean and (b) median improvement in $T_{90}^{\pm}$ relative to the experimentally measured response time without prediction across tested concentrations.

To assess the reliability of the LSTM ensemble-based uncertainty estimates, we quantitatively evaluated their calibration on the test set using empirical coverage probabilities (Fig. 5). The calibration curve demonstrates reasonable agreement with perfect calibration across a broad range of coverage levels, with a slight systematic bias towards under-confidence particularly at intermediate nominal coverage levels. This deviation diminishes toward both lower and higher coverage levels. Importantly, in the high-coverage regime most relevant for deployment (where predictive intervals are chosen to achieve nominal coverage probabilities of 90% or greater) empirical and nominal coverage are in near-perfect agreement.

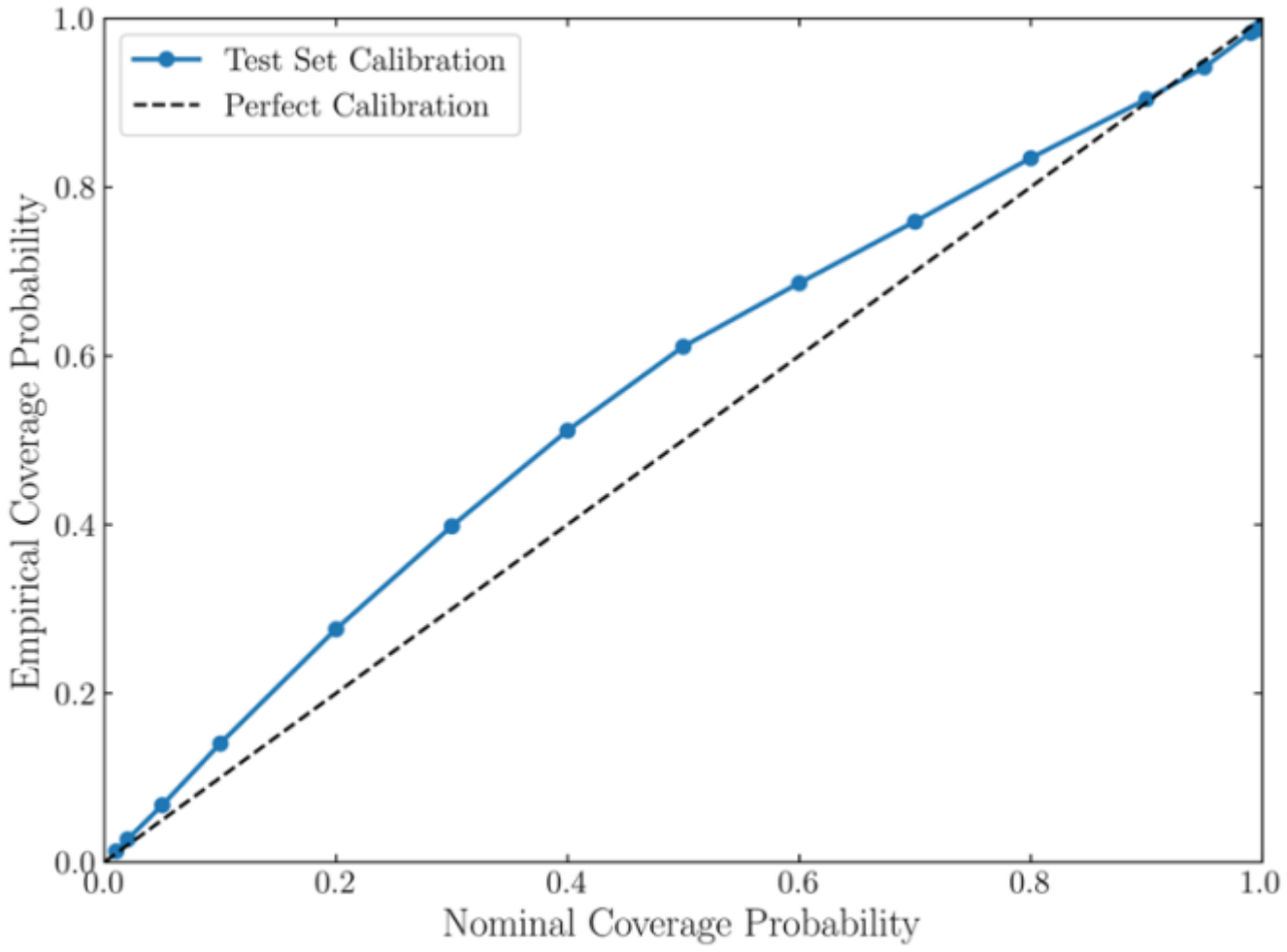

Fig. 5. Empirical coverage probability of ensemble prediction intervals as a function of nominal coverage probability, computed by pooling predictions across all test set samples. The dashed line indicates perfect calibration.

This demonstrates that when users request predictive intervals intended to contain the true equilibrium sensor response with high probability, the resulting uncertainty estimates are highly accurate despite substantial variation in signal-to-noise ratio across experimental conditions (Fig. 1, 2). In safety-critical deployment scenarios such as biosensing applications where predictions may inform downstream clinical decisions, well-calibrated uncertainty estimates are essential, as they enable principled thresholding, rejection of low-confidence predictions, and risk-aware decision making, supporting early yet trustworthy sensor readout without compromising reliability.

While the observed calibration is accurate on data drawn from the same experimental distribution as training, these results do not address performance under distributional shift. Future work will examine calibration robustness under shifted operating conditions and investigate approaches to improve calibration across the full range of coverage probabilities, including alternative uncertainty quantification methods such as adversarial training[30].

Further improvement in response time could be achieved with larger or augmented dataset. To this end, the utility of finite element method (FEM) modeling[16], [38] to provide a physics informed simulated dataset was explored. However, both transfer learning, and a single augmented training dataset containing experimental examples as well as FEM simulations, led to decreased performance rather than higher accuracy. This is likely a consequence of the complexity of the system, which cannot be captured using simulations in a sufficiently accurate, timely, and computationally feasible manner. More detailed discussion can be found in Appendix C. Additionally, other augmentation strategies such as interpolation could also enhance performance.

Finally, we note that while there are a wide range of models that can, in principle, be applied to time series forecasting problems, many are not well suited for this specific task due to the specific characteristics of biosensor response data. For example, approaches such as autoregressive integrated moving average (ARIMA) models and Kalman filters typically assume stationarity (perhaps with superimposed seasonality) and rely on linear or weakly-nonlinear state-space representations. In contrast, biosensor responses are strongly non-stationary, monotonic, with unknown and concentration-dependent asymptotic values, and are observed only once per assay without repeated steady-state measurements. These properties violate key assumptions required for reliable parameter estimation and extrapolation in such models, particularly when predictions must be made from limited early-time data. Additionally, physics-based models, while presenting an alternative approach for real-time extrapolation of equilibrium response, brings significant challenges in this context. In the absence of closed-form analytical solutions, physics-based extrapolation requires computationally intensive numerical simulations and ill-posed inverse parameter estimation, which becomes increasingly intractable when only early-time measurements are available. Collectively, these constraints limit the practicality of physics-based approaches for rapid, real-time prediction with robust uncertainty quantification. Nevertheless, such models provide valuable insight into mass transport and binding processes, play an essential role in sensor design and interpretation, and show some promise for the generation of synthetic data to augment experimental datasets (Appendix B).[59] The superior performance of LSTM ensembles is consistent with their explicit memory cell architecture, which facilitates retention of long-range temporal information and integration of informative early-time dynamics while attenuating noise. This architectural property is particularly advantageous for non-stationary biosensor responses where accurate prediction requires inferring a latent equilibrium state from incomplete observations.

## IV. Conclusion

In summary, we have demonstrated that ensembles of LSTM networks can reduce biosensor response time by predicting the ultimate equilibrium response, alongside a well calibrated measure of uncertainty, using a small quantity of initial time series sensor response data. The key advantages of the approach presented in this work are: i) indication of the confidence of the model's predictions which lays a crucial foundation for important safety-related decisions; ii) no requirement for prior domain knowledge of sensor response kinetics or manual feature engineering; iii) relatively high noise immunity learned by the model; iv) potential for generalization across different concentrations, assays, and sensor platforms due to the data-driven approach and the model's ability to learn temporal features directly from experimental data; v) scalability and high throughput data collection methodology enabling large datasets; vi) applicability in tandem with other response time reduction strategies. Through application to an experimental dataset collected by carrying out a BSA assay using PSi optical biosensors, for which mass transport and consequently response time is a significant limitation, a mean and median factor of improvement of 18.6 and 5.1, respectively, was realized. For the majority of sensors in the test dataset, the LSTM networks accurately predicted the sensor response in 45 minutes or less while the average experiment required just under 4 hours to reach the equilibrium response. The demonstrated dramatic reduction in biosensor response time sets the stage for faster medical diagnostics, which has potential to significantly improve patient-related outcomes and healthcare access (specifically in resource poor areas where scalable and rapid diagnostic tools would dramatically improve access to timely medical care), and earlier warning of toxins and harmful biomolecules in food and the environment.

## Appendix

### A. Validation of Stepper Motor Positioning Stability

To assess the positional repeatability of the stepper motor used for sequential optical spectra measurement of multiple PSi sensors, a dedicated control experiment was performed to quantify the maximum lateral displacement of the illumination spot over the course of an experiment. This is important to quantify since any drift in position or cumulative error could introduce spectral shifts that would confound adsorption induced changes.

A PSi calibration sample exhibiting a well-defined, monotonic gradient in optical thickness was prepared for this purpose. The sample was obtained from the periphery of a 1 $cm^2$ electrochemically etched PSi film, where a spatially uniform thickness gradient arises from edge effects in the etching cell. The local spatial gradient in optical thickness was characterized using a micrometer translation stage, enabling conversion of a measured spectral shift into an equivalent displacement.

The calibration sample was mounted such that the optical thickness gradient lay parallel to the direction of motion of the stepper motor, ensuring that any positional error along the motor trajectory would produce a measurable shift in the reflectance spectrum. The stepper motor was then operated under identical conditions to those used in all other sensing experiments, repeatedly stepping between predefined positions and returning to the same nominal location while reflectance spectra were continuously acquired over a 13 hour period.

The maximum observed variation in the reflectance peak position over the full 13 hour duration of the experiment corresponded to a lateral displacement of 90 μm. Over this length scale, the PSi sensors used in this work exhibited sufficient spatial uniformity that such displacement represents a negligible source of error relative to adsorption-induced spectral shifts. These results confirm that stepper motor positioning does not contribute appreciably to measurement variability in the reported experiments.

### B. Effect of Ensemble Size on Performance

Fig. B1 shows the improvement in the mean and median response time reduction across the validation dataset as ensemble size, or number of base learners, is increased. The improvement in performance can be seen to rise quickly, before beginning to plateau at approximately 15 base learners.

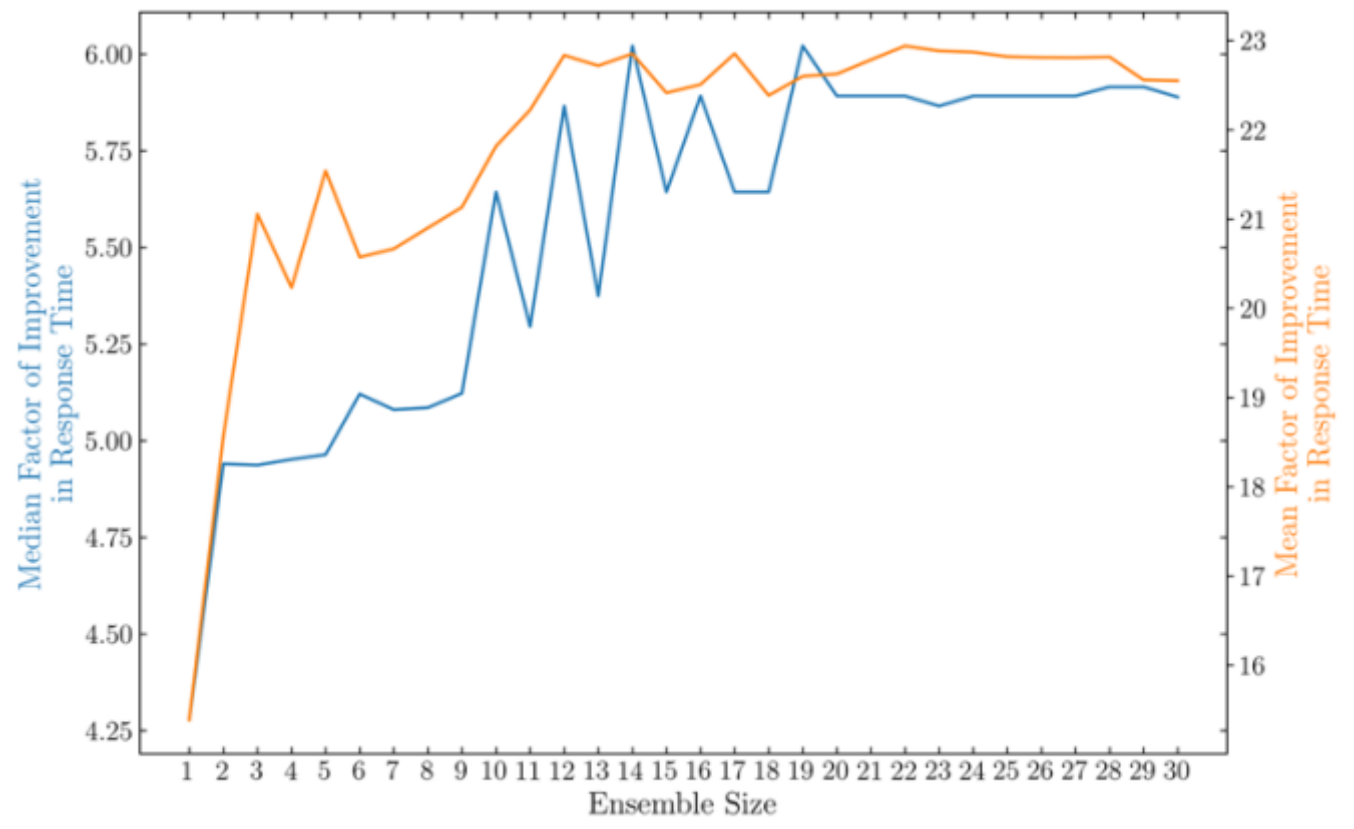

Fig. B1. Mean and median factor of improvement in biosensor response time realized when using ensembles of LSTM networks composed of different numbers of base learners.

## C. Transfer Learning with Simulated Data

A simulated dataset was compiled by fitting a 1D finite element method (FEM) model to each experimental sensor response and extracting best-fit physical parameters. These parameters were modelled with a multi-variate Gaussian distribution, new realistic sets of parameters were obtained by sampling from this distribution, and sensor time series responses were simulated using these parameters as input. The FEM model, presented in Ref 38, captures the mass transfer and binding kinetics in porous materials, assuming uniform cylindrical pores.

The physical input parameters fall into two categories, those fixed by the experimental conditions including BSA concentration, and those which were computed using gradient

TABLE CI
PARAMETERS FOR FEM MODEL

| Parameter | Range |
|---|---|
| Association Rate Constant $(M^{-1} \cdot s^{-1})$ | $10^1$ - $10^4$ |
| Dissociation Rate Constant $(s^{-1})$ | $10^{-5}$ - $10^0$ |
| Surface Binding Site Concentration $(mol \cdot m^{-2})$ | $10^{-8}$ - $10^{-4}$ |
| Pore Diameter $(nm)$ | 15 - 35 |
| Bulk Diffusion Coefficient $(m^2 \cdot s^{-1})$ | $10^{-11}$ - $10^{-9}$ |
| BSA Hydrodynamic Radius $(nm)$ | 4 - 6 |

Effective physical input parameters of the FEM model, alongside the range of values explored by gradient based optimization, to minimize mean squared error between simulated and experimental sensor response data.

optimization methods, alongside the maximum and minimum values between which each parameter was constrained.

While all of the parameters in Table C1 are measurable, the goal is to find the parameter values that will lead to simulations that match real sensor response data the closest. This corresponds to the best estimate of parameters that minimize mean squared error between the simulated and experimental time series sensor response, accounting for the inaccuracies introduced by assumptions of uniform cylindrical pores, and ignoring the distribution of pore sizes, non-cylindrical shape, and pore branching. The set of extracted parameters are visualized using linear discriminant analysis to reduce dimensionality in Fig. C1, which reveals the extent to which parameters can be separated by analyte concentration.

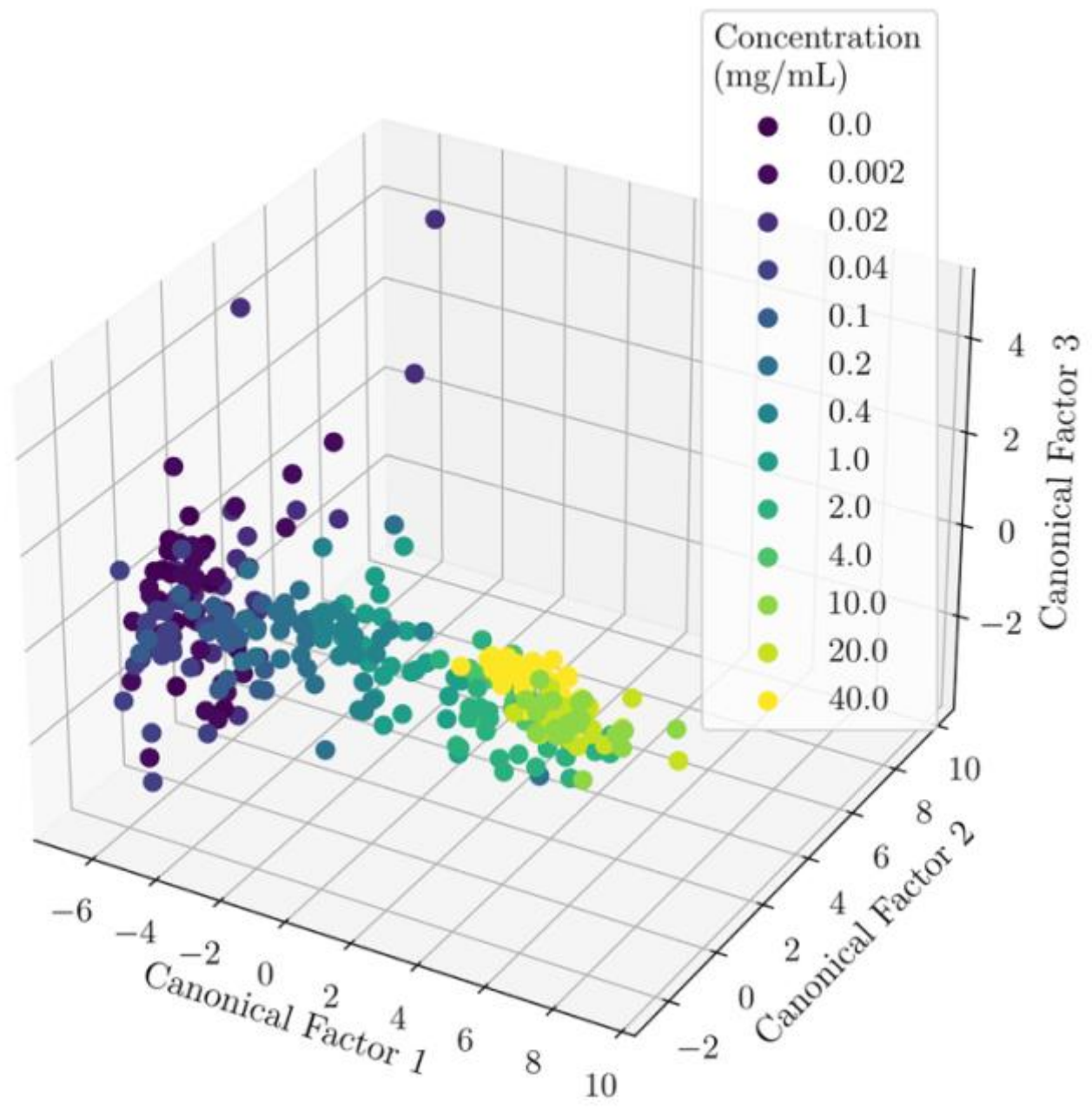

Fig. C1. Dimensionality reduced visualization of the set of physical parameters extracted by fitting an FEM model to the experimental sensor response time series data.

Fig. C2 shows a sample of the simulated dataset equally distributed across the range of experimental BSA concentrations explored, created by sampling from the distribution of extracted physical parameters (Fig. C1), simulating the sensor response using these parameters as input to the FEM model, and superimposing white Gaussian noise achieving S/N level within the range of experimentally derived values. The qualitative similarities between simulated and experimental sensor response curves can be seen by comparing Fig. C2 and Fig. 1b.

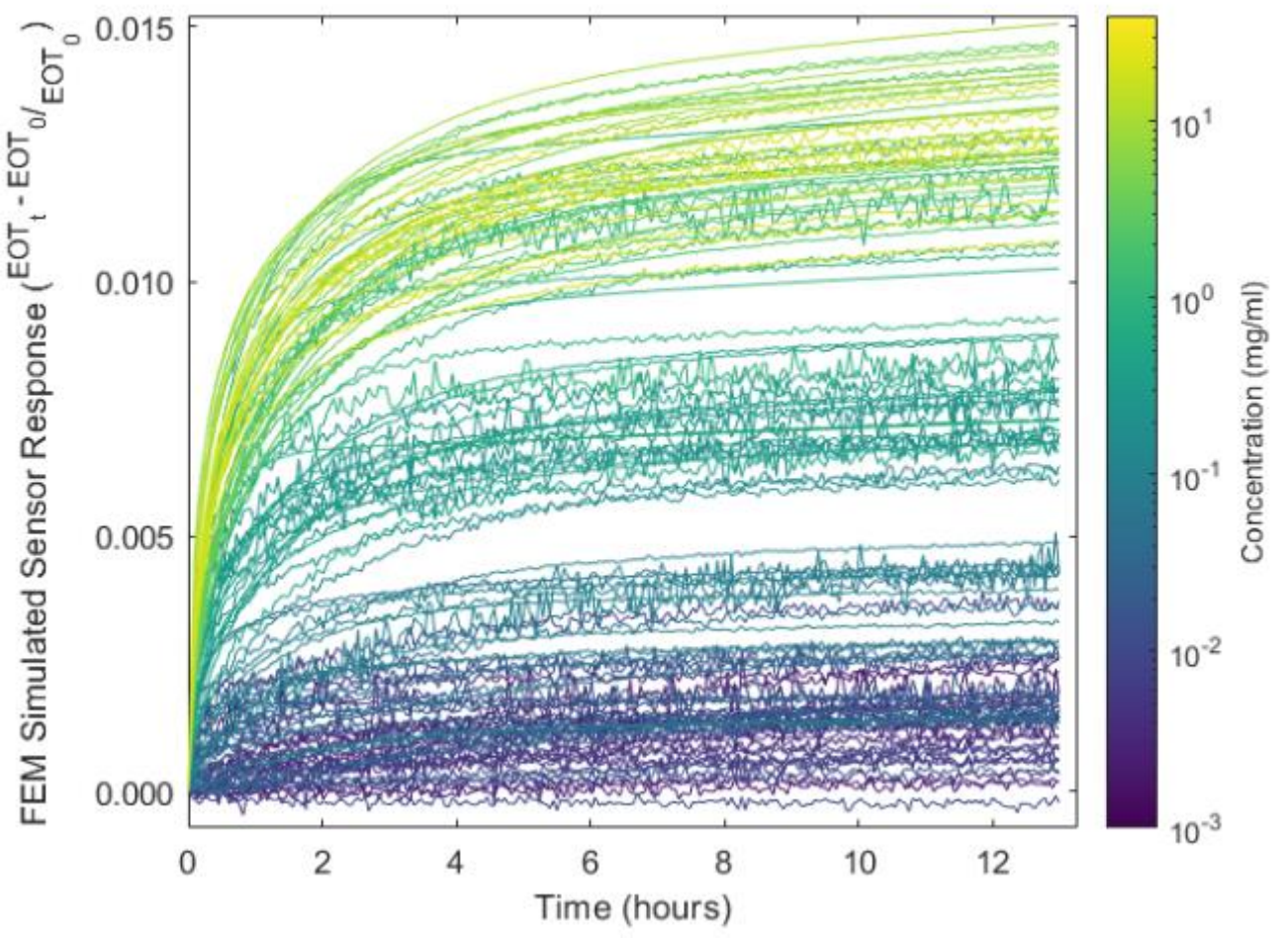

Fig. C2. Representative sample of the simulated data set using physics-based FEM model, extracted parameters from experimental data, and superimposing white Gaussian noise.

Unfortunately, despite the huge increase in dataset size that the simulated data enables (387 to >10,000 examples), neither transfer learning on a large simulated dataset and fine-tuning on real experimental data, or training on a large augmented dataset consisting of both real and experimental data, improved prediction accuracy significantly, so simulated data was not used to train the final ensemble. This is expected to be due to the simulated data lying significantly outside the distribution of experimental data, and could be addressed by more accurate modelling of a) the sensor responses with a more complex FEM model, and b) the statistical distribution of best-fit physical parameters using copulas or generative methods as opposed to a multi-variate Gaussian assumption.

## Acknowledgment

Porous silicon wafers were fabricated and diced in the Vanderbilt Institute of Nanoscale Science and Engineering. The authors thank Catie Chang for useful discussions contributing to this work.

## References

[1] V. Gubala, L. F. Harris, A. J. Ricco, M. X. Tan, and D. E. Williams, "Point of care diagnostics: Status and future," *Anal. Chem.*, vol. 84, no. 2, pp. 487–515, 2012, doi: 10.1021/ac2030199.

[2] F. Ko and S. J. Drews, "The impact of commercial rapid respiratory virus diagnostic tests on patient outcomes and health system utilization," *Expert Rev. Mol. Diagn.*, vol. 17, no. 10, pp. 917–931, 2017, doi: 10.1080/14737159.2017.1372195.

[3] J. Barenfanger, C. Drake, N. Leon, T. Mueller, and T. Troutt, "Clinical and financial benefits of rapid detection of respiratory viruses: An outcomes study," *J. Clin. Microbiol.*, vol. 38, no. 8, pp. 2824–2828, 2000, doi: 10.1128/jcm.38.8.2824-2828.2000.

[4] B. B. Rogers *et al.*, "Impact of a rapid respiratory panel test on patient outcomes," *Arch. Pathol. Lab. Med.*, vol. 139, no. 5, pp. 636–641, 2015, doi: 10.5858/arpa.2014-0257-OA.

[5] J. Wang and H. Yue, "Food safety pre-warning system based on data mining for a sustainable food supply chain," *Food Control*, vol. 73, pp. 223–229, 2017, doi: 10.1016/j.foodcont.2016.09.048.

[6] Z. Q. Geng, S. S. Zhao, G. C. Tao, and Y. M. Han, "Early warning modeling and analysis based on analytic hierarchy process integrated extreme learning machine (AHP-ELM): Application to food safety," *Food Control*, vol. 78, pp. 33–42, 2017, doi: 10.1016/j.foodcont.2017.02.045.

[7] J. Hellou, "Behavioural ecotoxicology, an 'early warning' signal to assess environmental quality," *Environmental Science and Pollution Research*, vol. 18, no. 1, pp. 1–11, 2011, doi: 10.1007/s11356-010-0367-2.

[8] F. Long, A. Zhu, and H. Shi, "Recent advances in optical biosensors for environmental monitoring and early warning," *Sensors*, vol. 13, no. 10, pp. 13928–13948, 2013, doi: 10.3390/s131013928.

[9] Z. Li *et al.*, "Hydrogen Gas Sensor Based on Mesoporous $In_2O_3$ with Fast Response/Recovery and ppb Level Detection Limit," *Int. J. Hydrogen Energy*, vol. 43, no. 50, pp. 22746–22755, 2018, doi: 10.1016/j.ijhydene.2018.10.101.

[10] L. T. Duy *et al.*, "Flexible Transparent Reduced Graphene Oxide Sensor Coupled with Organic Dye Molecules for Rapid Dual-Mode Ammonia Gas Detection," *Adv. Funct. Mater.*, vol. 26, no. 24, pp. 4329–4338, 2016, doi: 10.1002/adfm.201505477.

[11] X. Meng, M. Bi, Q. Xiao, and W. Gao, "Ultra-fast response and highly selectivity hydrogen gas sensor based on Pd/$SnO_2$ nanoparticles," *Int. J. Hydrogen Energy*, vol. 47, no. 5, pp. 3157–3169, 2022, doi: 10.1016/j.ijhydene.2021.10.201.

[12] M. A. Gertz, "Acute hyperviscosity: Syndromes and management," *Blood*, vol. 132, no. 13, pp. 1379–1385, 2018, doi: 10.1182/blood-2018-06-846816.

[13] S. K. Kumar *et al.*, "Waldenström Macroglobulinemia/ Lymphoplasmacytic Lymphoma, Version 2.2024," *JNCCN Journal of the National Comprehensive Cancer Network*, vol. 22, no. 1, pp. 1–19, 2024, doi: 10.6004/jnccn.2024.0001.

[14] I. Jokić, Z. Djurić, K. Radulović, M. Frantlović, G. V. Milovanović, and P. M. Krstajić, "Stochastic time response and ultimate noise performance of adsorption-based microfluidic biosensors," *Biosensors (Basel).*, vol. 11, no. 6, p. 194, 2021, doi: 10.3390/bios11060194.

[15] Y. Zhao, G. Gaur, R. L. Mernaugh, P. E. Laibinis, and S. M. Weiss, "Comparative Kinetic Analysis of Closed-Ended and Open-Ended Porous Sensors," *Nanoscale Res. Lett.*, vol. 11, no. 395, pp. 1–9, Dec. 2016, doi: 10.1186/s11671-016-1614-3.

[16] Y. Zhao, G. Gaur, S. T. Retterer, P. E. Laibinis, and S. M. Weiss, "Flow-through porous silicon membranes for real-time label-free biosensing," *Anal. Chem.*, vol. 88, no. 22, pp. 10940–10948, 2016, doi: 10.1021/acs.analchem.6b02521.

[17] M. R. Leyden *et al.*, "Increasing the detection speed of an all-electronic real-time biosensor," *Lab Chip*, vol. 12, no. 5, pp. 954–959, 2012, doi: 10.1039/c2lc21020g.

[18] J. Lu, I. Do, L. T. Drzal, R. M. Worden, and I. Lee, "Nanometal-decorated exfoliated graphite nanoplatelet based glucose biosensors with high sensitivity and fast response," *ACS Nano*, vol. 2, no. 9, pp. 1825–1832, 2008, doi: 10.1021/nn800244k.

[19] L. Soleymani and F. Li, "Mechanistic Challenges and Advantages of Biosensor Miniaturization into the Nanoscale," *ACS Sens.*, vol. 2, no. 4, pp. 458–467, 2017, doi: 10.1021/acssensors.7b00069.

[20] J. Zhang *et al.*, "Reduction of Biosensor False Responses and Time Delay Using Dynamic Response and Theory-Guided Machine Learning," *ACS Sens.*, vol. 8, no. 11, pp. 4079–4090, 2023, doi: 10.1021/acssensors.3c01258.

[21] M. K. Muezzinoglu *et al.*, “Acceleration of chemo-sensory information processing using transient features,” *Sens. Actuators B Chem.*, vol. 137, no. 2, pp. 507–512, 2009, doi: 10.1016/j.snb.2008.10.065.

[22] M. Siadat, H. Sambemana, and M. Lumbrerasa, “New transient feature for metal oxide gas sensor response processing,” *Procedia Eng.*, vol. 47, pp. 52–55, 2012, doi: 10.1016/j.proeng.2012.09.082.

[23] S. Kanaparthi and S. G. Singh, “Reduction of the Measurement Time of a Chemiresistive Gas Sensor Using Transient Analysis and the Cantor Pairing Function,” *ACS Measurement Science Au*, vol. 2, no. 2, pp. 113–119, 2022, doi: 10.1021/acsmeasuresciau.1c00043.

[24] D. Ziaian *et al.*, “Improving Systems Dynamics by Means of Advanced Signal Processing – Mathematical , Laboratorial and Clinical Evaluation of Propofol Monitoring in Breathing Gas,” *Sensors & Transducers*, vol. 193, no. 10, pp. 145–153, 2015, [Online]. Available: http://www.sensorsportal.comhttp//www.sensorsportal.com/HTML/DIGEST/P_2747.htm

[25] D. L. Osorio-Arrieta *et al.*, “Reduction of the measurement time by the prediction of the steady-state response for quartz crystal microbalance gas sensors,” *Sensors*, vol. 18, no. 8, 2018, doi: 10.3390/s18082475.

[26] J. Fonollosa, S. Sheik, R. Huerta, and S. Marco, “Reservoir computing compensates slow response of chemosensor arrays exposed to fast varying gas concentrations in continuous monitoring,” *Sens. Actuators B Chem.*, vol. 215, pp. 618–629, 2015, doi: 10.1016/j.snb.2015.03.028.

[27] Q. Zhang, S. Li, W. Tang, and X. Guo, “Fast Measurement with Chemical Sensors Based on Sliding Window Sampling and Mixed-Feature Extraction,” *IEEE Sens. J.*, vol. 20, no. 15, pp. 8740–8745, 2020, doi: 10.1109/JSEN.2020.2985034.

[28] J. Lee, H. Hong, J. M. Song, and E. Yeom, “Neural network ensemble model for prediction of erythrocyte sedimentation rate (ESR) using partial least squares regression,” *Sci. Rep.*, vol. 12, no. 1, pp. 1–13, 2022, doi: 10.1038/s41598-022-23174-0.

[29] M. Abdar, F. Pourpanah, S. Hussain, D. Rezazadegan, and L. Liu, “A review of uncertainty quantification in deep learning: Techniques , applications and challenges,” *Information Fusion*, vol. 76, pp. 243–297, 2021, doi: 10.1016/j.inffus.2021.05.008.

[30] B. Lakshminarayanan, A. Pritzel, and C. Blundell, “Simple and scalable predictive uncertainty estimation using deep ensembles,” in *Proceedings of the 31st International Conference on Neural Information Processing Systems (NIPS’17)*, 2017, pp. 6403–6414. doi: 10.1109/34.58871.

[31] G. Rong, A. Najmaie, J. E. Sipe, and S. M. Weiss, “Nanoscale porous silicon waveguide for label-free DNA sensing,” vol. 23, pp. 1572–1576, 2008, doi: 10.1016/j.bios.2008.01.017.

[32] O. Syshchyk, V. A. Skryshevsky, O. O. Soldatkin, and A. P. Soldatkin, “Enzyme biosensor systems based on porous silicon photoluminescence for detection of glucose, urea and heavy metals,” *Biosensors and Bioelectronic*, vol. 66, pp. 89–94, 2015, doi: 10.1016/j.bios.2014.10.075.

[33] N. Massad-Ivanir *et al.*, “Porous Silicon-Based Biosensors: Towards Real-Time Optical Detection of Target Bacteria in the Food Industry,” *Sci. Rep.*, vol. 6, Nov. 2016, doi: 10.1038/srep38099.

[34] V. S. Y. Lin, K. Motesharei, K. P. S. Dancil, M. J. Sailor, and M. R. Ghadiri, “A porous silicon-based optical interferometric biosensor,” *Science (1979).*, vol. 278, no. 5339, pp. 840–843, 1997, doi: 10.1126/science.278.5339.840.

[35] A. Jane, R. Dronov, A. Hodges, and N. H. Voelcker, “Porous silicon biosensors on the advance,” *Trends Biotechnol.*, vol. 27, no. 4, pp. 230–239, Apr. 2009, doi: 10.1016/j.tibtech.2008.12.004.

[36] R. Herino, G. Bomchil, K. Barla, C. Bertrand, and J. L. Ginoux, “Porosity and Pore Size Distributions of Porous Silicon Layers,” *J. Electrochem. Soc*, vol. 134, no. 8, pp. 1994–2000, Aug. 1987, doi: 10.1149/1.2100805.

[37] M. Boehringer, H. Artmann, and K. Witt, “Porous Silicon in a Semiconductor Manufacturing Environment,” *Journal of Microelectromechanical Systems*, vol. 21, no. 6, pp. 1375–1381, Dec. 2012, doi: 10.1109/JMEMS.2012.2205900.

[38] S. Arshavsky Graham, E. Boyko, R. Salama, and E. Segal, “Mass Transfer Limitations of Porous Silicon-Based Biosensors for Protein Detection,” *ACS Sens.*, vol. 5, no. 10, pp. 3058–3069, 2020, doi: 10.1021/acssensors.0c00670.

[39] S. J. Ward, T. Cao, X. Zhou, C. Chang, and S. M. Weiss, “Protein Identification and Quantification Using Porous Silicon Arrays, Optical Measurements , and Machine Learning,” *Biosensors (Basel).*, vol. 13, no. 879, pp. 1–12, 2023, doi: https://doi.org/10.3390/bios1309087.

[40] M. J. Sailor, *Porous Silicon in Practice: Preparation, Characterization and Applications*, 1st ed. Weinheim, Germany: Wiley-VCH, 2012. doi: 10.1002/9783527641901.

[41] C. Pacholski, M. Sartor, M. J. Sailor, F. Cunin, and G. M. Miskelly, “Biosensing using porous silicon double-layer interferometers: Reflective interferometric fourier transform spectroscopy,” *J. Am. Chem. Soc.*, vol. 127, no. 33, pp. 11636–11645, 2005, doi: 10.1021/ja0511671.

[42] I. Rea, E. Orabona, A. Lamberti, I. Rendina, and L. De Stefano, “A microfluidics assisted porous silicon array for optical label-free biochemical sensing,” *Biomicrofluidics*, vol. 5, no. 3, pp. 1–10, 2011, doi: 10.1063/1.3626008.

[43] S. Arshavsky-Graham, A. Enders, S. Ackerman, J. Bahnemann, and E. Segal, “3D-printed microfluidics integrated with optical nanostructured porous aptasensors for protein detection,” *Microchimica Acta*, vol. 188, no. 67, pp. 1–12, 2021, doi: 10.1007/s00604-021-04725-0.

[44] D. J. C. MacKay, “Probable networks and plausible predictions - a review of practical Bayesian methods

for supervised neural networks," *Network: computation in neural systems*, vol. 6, no. 3, pp. 469–505, 1995, doi: https://doi.org/10.1088/0954-898X_6_3_011.

[45] N. Metropolis, A. W. Rosenbluth, M. N. Rosenbluth, A. H. Teller, and E. Teller, "Equation of State Calculations by Fast Computing Machines," *Journal of Chemical Physics*, vol. 21, no. 6, pp. 1087–1092, 1953, doi: https://doi.org/10.1063/1.1699114.

[46] R. Zhao, R. Yan, J. Wang, and K. Mao, "Learning to monitor machine health with convolutional Bi-directional LSTM networks," *Sensors*, vol. 17, no. 273, pp. 1–18, 2017, doi: 10.3390/s17020273.

[47] F. Weninger *et al.*, "Speech Enhancement with LSTM Recurrent Neural Networks and its Application to Noise-Robust ASR," in *Brain Research Bulletin*, vol. 53, no. 2, 2015, pp. 91–99. doi: 10.1007/978-3-319-22482-4_11.

[48] X. H. Le, H. V. Ho, G. Lee, and S. Jung, "Application of Long Short-Term Memory (LSTM) neural network for flood forecasting," *Water (Basel).*, vol. 11, no. 7, 2019, doi: 10.3390/w11071387.

[49] C. Zhou, C. Sun, Z. Liu, and F. C. M. Lau, "A C-LSTM Neural Network for Text Classification," *ArXiv*, vol. abs/1511.0, 2015.

[50] S. Hochreiter and J. Schmidhuber, "Long Short-Term Memory," *Neural Comput.*, vol. 9, no. 8, pp. 1735–1780, Nov. 1997, doi: 10.1162/neco.1997.9.8.1735.

[51] F. Chollet and others, "Keras," 2015, *GitHub*.

[52] L. K. Hansen and P. Salamon, "Neural network ensembles," *IEEE Trans. Pattern Anal. Mach. Intell.*, vol. 12, no. 10, pp. 993–1001, 1990, doi: 10.1109/34.58871.

[53] D. P. Kingma and J. L. Ba, "Adam: A method for stochastic optimization," *ArXiv*, vol. abs/1412.6, pp. 1–15, 2015.

[54] O. Redlich and D. L. Peterson, "A Useful Adsorption Isotherm," *J. Phys. Chem.*, vol. 63, no. 6, p. 1024, Jun. 1959, doi: 10.1021/j150576a611.

[55] S. J. Ward, R. Layouni, S. Arshavsky-Graham, E. Segal, and S. M. Weiss, "Morlet Wavelet Filtering and Phase Analysis to Reduce the Limit of Detection for Thin Film Optical Biosensors," *ACS Sens.*, vol. 6, no. 8, pp. 2967–2978, 2021, doi: 10.1021/acssensors.1c00787.

[56] R. Moretta, L. De Stefano, M. Terracciano, and I. Rea, "Porous Silicon Optical Devices: Recent Advances in Biosensing Applications," *Sensors*, vol. 21, no. 4, p. 1336, 2021, doi: 10.3390/s21041336.

[57] C. K. Tsang, T. L. Kelly, M. J. Sailor, and Y. Y. Li, "Highly Stable Porous Silicon–Carbon Composites as Label-Free Optical Biosensors," *ACS Nano*, vol. 6, no. 12, pp. 10546–10554, 2012, doi: 10.1021/nn304131d.

[58] F. Wang, T. J. Barnes, and C. A. Prestidge, "Controlling and predicting the dissolution kinetics of thermally oxidised mesoporous silicon particles: Towards improved drug delivery," *Pharmaceutics*, vol. 11, no. 12, Dec. 2019, doi: 10.3390/pharmaceutics11120634.